\documentclass[letterpaper, 10 pt, conference]{ieeeconf}  

\IEEEoverridecommandlockouts                              

\usepackage{graphicx}
\usepackage{subcaption}
\usepackage{multirow}
\usepackage{amssymb}
\title{\LARGE \bf
Beyond Object Categories: Multi-Attribute Reference Understanding for Visual Grounding
}

\author{
Hao Guo, Jianfei Zhu, Wei Fan, Chunzhi Yi, ~\IEEEmembership{Member,~IEEE}, Feng Jiang, ~\IEEEmembership{Senior Member,~IEEE}
}

\begin{document}

\maketitle
\thispagestyle{empty}
\pagestyle{empty}

\begin{abstract}

Referring expression comprehension (REC) aims at achieving object localization based on natural language descriptions. However, existing REC approaches are constrained by object category descriptions and single-attribute intention descriptions, hindering their application in real-world scenarios. In natural human-robot interactions, users often express their desires through individual states and intentions, accompanied by guiding gestures, rather than detailed object descriptions. To address this challenge, we propose Multi-ref EC, a novel task framework that integrates state descriptions, derived intentions, and embodied gestures to locate target objects. We introduce the State-Intention-Gesture Attributes Reference (SIGAR) dataset, which combines state and intention expressions with embodied references. Through extensive experiments with various baseline models on SIGAR, we demonstrate that properly ordered multi-attribute references contribute to improved localization performance, revealing that single-attribute reference is insufficient for natural human-robot interaction scenarios. Our findings underscore the importance of multi-attribute reference expressions in advancing visual-language understanding.

\end{abstract}

\section{INTRODUCTION}

The advancement of embodied intelligence has driven human-robot interaction (HRI) closer to natural human communication \cite{tang2020bootstrapping}. A crucial capability for embodied agents is to locate objects based on natural language expressions. In real-world scenarios, users often express their desires for objects implicitly through state expressions (e.g., "I just finished bathing" for towel) or derived intentions (e.g., "I would like to dry my body" for towel), frequently accompanied by guiding gestures \cite{herbort2016spatial,herbort2018point}. This complex interaction pattern requires agents to understand and reason about multi-attribute references for accurate object localization, presenting challenges beyond traditional referring expression comprehension (REC) tasks.

Two fundamental challenges emerge in multi-attribute reference understanding. First, implicit reference comprehension requires agents to understand users' potential needs rather than explicit object descriptions. While recent studies have explored intention-based visual grounding (VG) \cite{qu2024rio,wang2024beyond}, existing approaches predominantly focus on intention-level understanding, overlooking the crucial role of state descriptions in human expression. Understanding and reasoning about human states serves as a fundamental prerequisite for intelligent agents, as states not only reflect users' immediate needs but also provide crucial context for interpreting their intentions, thereby enabling more accurate and contextually appropriate assistance in real-world scenarios. Furthermore, the semantic gap between state expressions and target objects presents additional challenges, as it requires multi-step reasoning to bridge users' implicit needs with appropriate object selections.

The second fundamental challenge lies in embodied reference understanding, which demands visual perspective-taking (VPT) \cite{qiu2020human} capabilities, enabling agents to share perceptual fields with users. Although existing research has proposed methods like virtual touch-line perception \cite{li2023understanding}, these approaches often rely heavily on pre-annotated gesture keypoints and fixed geometric patterns, limiting their flexibility and generalization. Moreover, the heterogeneous nature of these references—embodied cues requiring extraction from visual features and implicit references encoded in natural language—poses significant challenges in multimodal fusion and cross-modal reasoning, making it crucial to develop effective integration strategies for robust model performance.

\begin{figure}[tb]
\centerline{\includegraphics[width=3.4in]{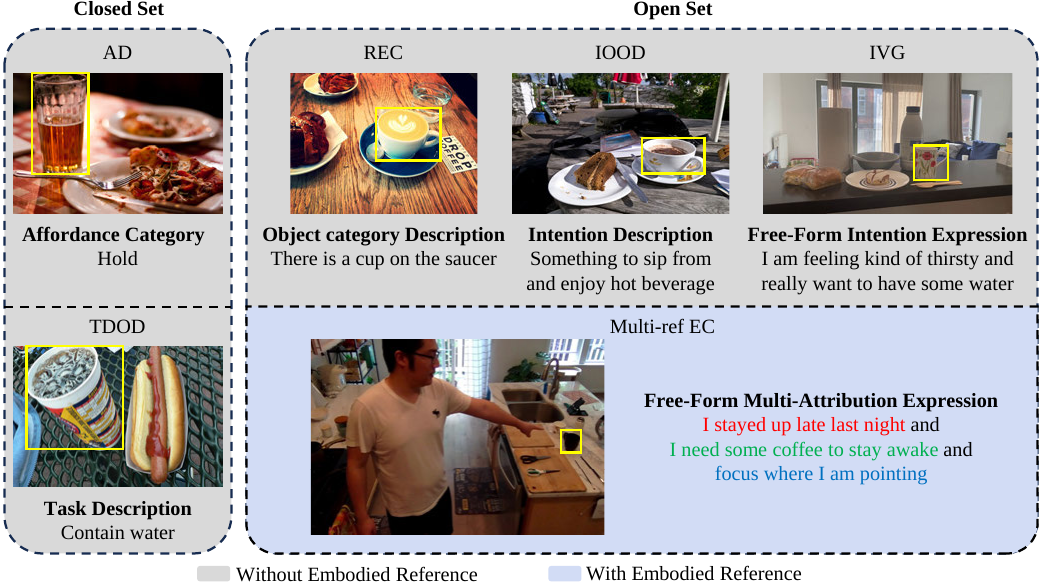}}
\caption{Comparison of visual grounding paradigms: Affordance Detection (AD), Task-Driven Object Detection (TDOD), Referring Expression Comprehension (REC), Intention-Oriented Object Detection (IOOD), Intention-driven VG (IVG), and our proposed Multi-ref EC. Red, green, and blue text in Multi-ref EC represent state, intention, and gesture references, respectively.}
\label{fig1}
\end{figure}

To address these limitations, we propose Multi-referring Expression Comprehension (Multi-ref EC) as shown in Fig. \ref{fig1}, extending traditional REC tasks to incorporate state-based intentions and embodied references for object localization. Our approach enables agents to understand and reason about multiple reference attributes simultaneously, bridging the gap between current technology and natural human communication patterns. We introduce the State-Intention-Gesture Attributes Reference (SIGAR) dataset, built upon YouRefIt \cite{chen2021yourefit}, with additional annotations for human states and intentions. This comprehensive dataset provides a solid foundation for developing and evaluating models that can handle complex multimodal references. We establish comprehensive baselines including combinatorial approaches, end-to-end models, and multimodal large language models, with the latter achieving state-of-the-art (SOTA) performance on our benchmark.
Our main contributions are threefold:

1) We introduce Multi-ref EC, a novel task framework that better aligns with natural interaction environments by incorporating multiple reference attributes. This framework advances the field beyond simple object-level understanding to complex multimodal reasoning.

2) We develop SIGAR, to the best of our knowledge, the first comprehensive benchmark for evaluating multimodal reference-based localization with free-form multi-attribute annotations. This dataset enables systematic study of how different reference types contribute to object localization.

3) We establish strong baselines for Multi-ref EC and conduct extensive experiments to analyze the effectiveness of different attribute combinations and their optimal organization patterns, providing valuable insights for future research in multi-attribute reference understanding.

\section{Related Work}

\subsection{Referring Expression Comprehension (REC)}

REC is an advanced VG task, aims to locate specific objects in images based on natural language descriptions. Previous REC tasks \cite{zhang2024vision,yao2024detclipv3,kamath2021mdetr} have focused on object category descriptions, and recently some studies have begun to explore non-directive descriptions such as intentions from the user's perspective. In particular, \cite{wang2024beyond} and \cite{qu2024rio} implemented VG tasks with different forms of intention-based text, and although these studies demonstrate the importance of locating an object based on an intention-based text, they were limited to textual descriptions of single attribute. In our task, we employ intent texts and state texts in linguistic information, while incorporating embodied references in non-linguistic information for understanding natural human expressions in multiple modalities and multiple attribute references.

\subsection{Vision-Language Complex Reasoning (VLCR)}

Real-world natural language interactions often involve complex texts with rich information rather than simple object-level statements. Traditional vision-language models excel in specific tasks \cite{li2022grounded, liu2023grounding}, they lack the generalized knowledge and natural language understanding required for VLCR. Recent works have leveraged large language models to bridge this gap \cite{pi2023detgpt}. Notable examples include RIO \cite{qu2024rio}, which introduced the IOOD dataset for object function description. IVG \cite{wang2024beyond} further advanced the field with the IntentionVG dataset, incorporating free-form intentional texts and first-person perspective multi-scene perception. However, users usually subconsciously describe subjective states to implicitly express their intentions, requiring the agents to understand and reason about the objects in the scenario that can satisfy the user's potential intentions.

\subsection{Embodied Reference Understanding}

Non-linguistic references, particularly pointing gestures, play a crucial role in human communication by providing precise direction information. The YouRefIt, introduced by \cite{chen2021yourefit} combines linguistic information with human pointing gestures, demonstrating gestural cues are as critical as language cues in understanding the embodied reference. \cite{li2023understanding} enhanced the localization performance with embodied reference through VTL-based pose understanding. Our research also introduces the pointing gesture as embodied reference to refine the target visual area, coupling with linguistic information to enhance the grounding performance.

\section{Dataset}

\subsection{Dataset Collection}

\begin{figure}[tb]
\centerline{\includegraphics[width=3.4in]{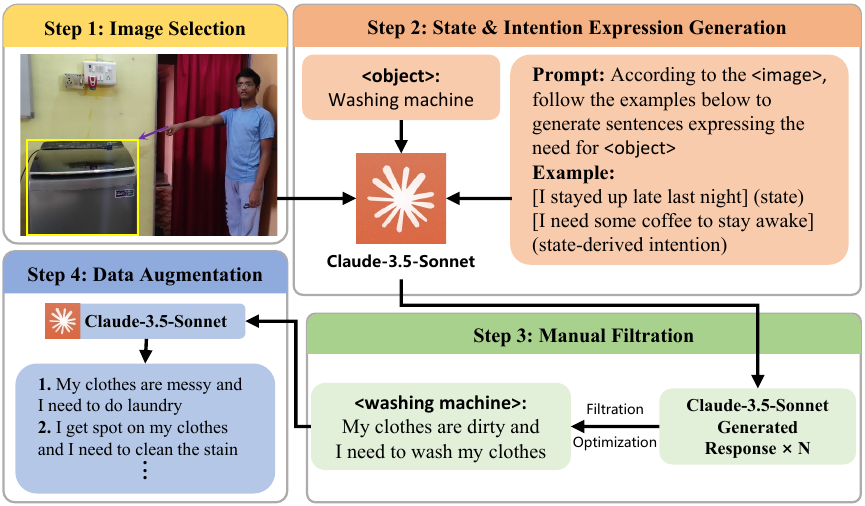}}
\caption{Illustration of the data collection for Multi-ref EC. The process begins with inheriting YouRefIt data, followed by generating state-intention drafts using Claude-3.5-sonnet with vision-language input. We then manually filter for well-matched expressions and apply data augmentation to create semantically equivalent variations.}
\label{fig2}
\end{figure}

Our dataset builds upon YouRefIt, with a two-phase collection process: base dataset development and data augmentation. The data collection process is shown in Fig. \ref{fig2}.

\paragraph{Base Dataset Development}
We leverage the YouRefIt dataset's existing annotations, including images, object category descriptions, and bounding box information. To understand given scenarios and generate corresponding expressions, we employ Claude-3.5-Sonnet \cite{claude2024sonnet}, a SOTA multimodal large language model (MLLM). Through carefully designed prompts with examples (i.e., "\textit{According to the \textless image\textgreater, generate sentences expressing the need for \textless object\textgreater. Example for cup: [I stayed up late last night] (state), [I need some coffee to stay awake] (state-derived intention)}"), we query the MLLM to generate first-person expressions reflecting user states and intentions in human-robot interactions. These generated texts undergo manual review to ensure contextual alignment with scene-specific object usage (e.g., distinguishing between towels in bathroom and rags in kitchen).

\paragraph{Data Augmentation}

To enhance model adaptation to the Multi-ref EC task, we perform text-based data augmentation using Claude-3.5-Sonnet. Unlike the base dataset generation, this phase focuses solely on language input, excluding visual components. We employ specialized prompts (i.e., “\textit{Based on the original annotation: \textless state\textgreater and \textless intention\textgreater, generate semantically equivalent variations while maintaining the same state-intention relationship.}”) to generate augmented texts. All augmented texts undergo manual filtering to ensure quality and relevance.

\subsection{Dataset Statistics}
\begin{figure}[htbp]
\centering
\begin{subfigure}[b]{0.32\columnwidth}
    \includegraphics[width=\textwidth]{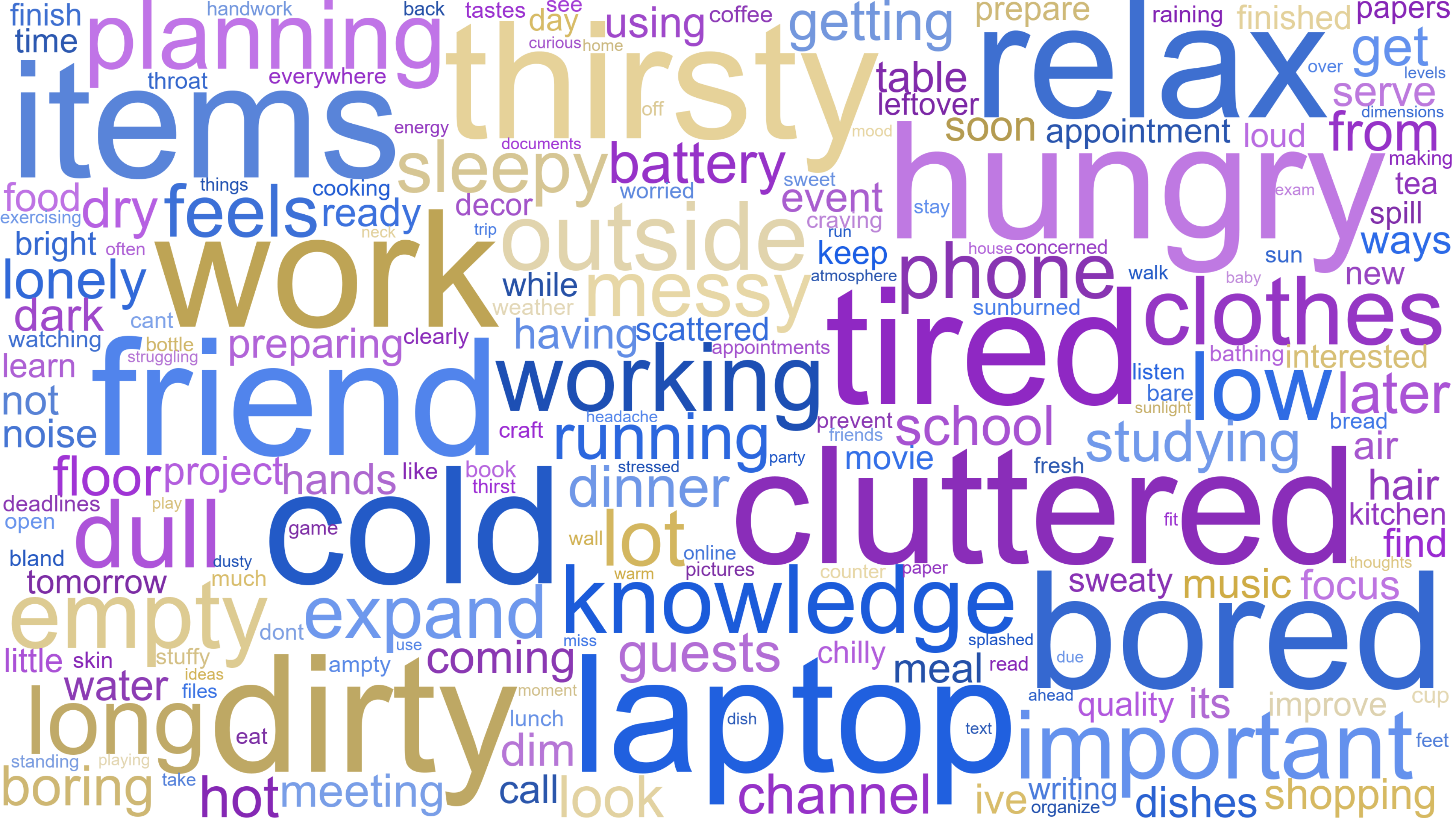}
    \caption{}
\end{subfigure}
\hfill
\begin{subfigure}[b]{0.32\columnwidth}
    \includegraphics[width=\textwidth]{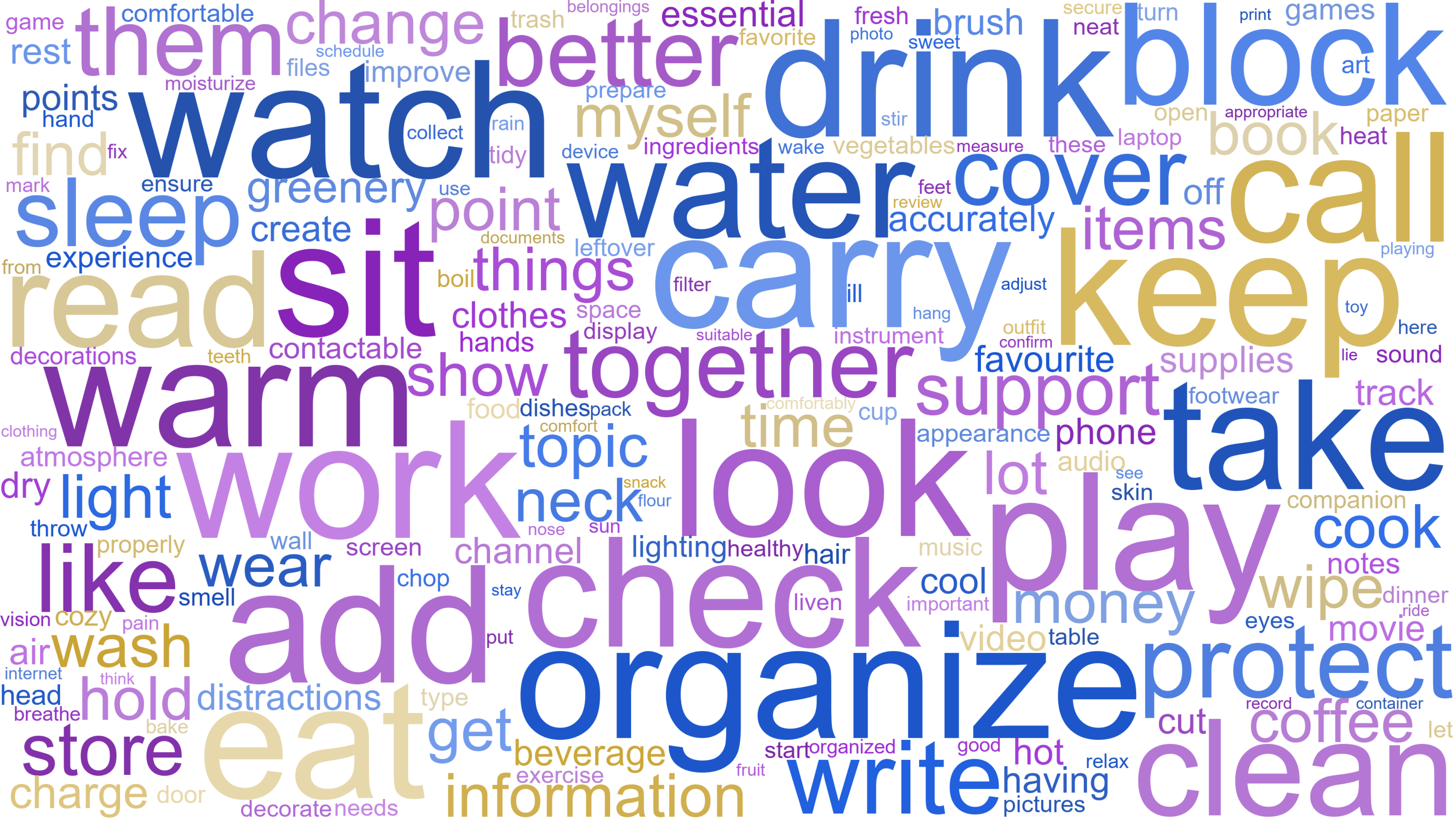}
    \caption{}
\end{subfigure}
\hfill
\begin{subfigure}[b]{0.32\columnwidth}
    \includegraphics[width=\textwidth]{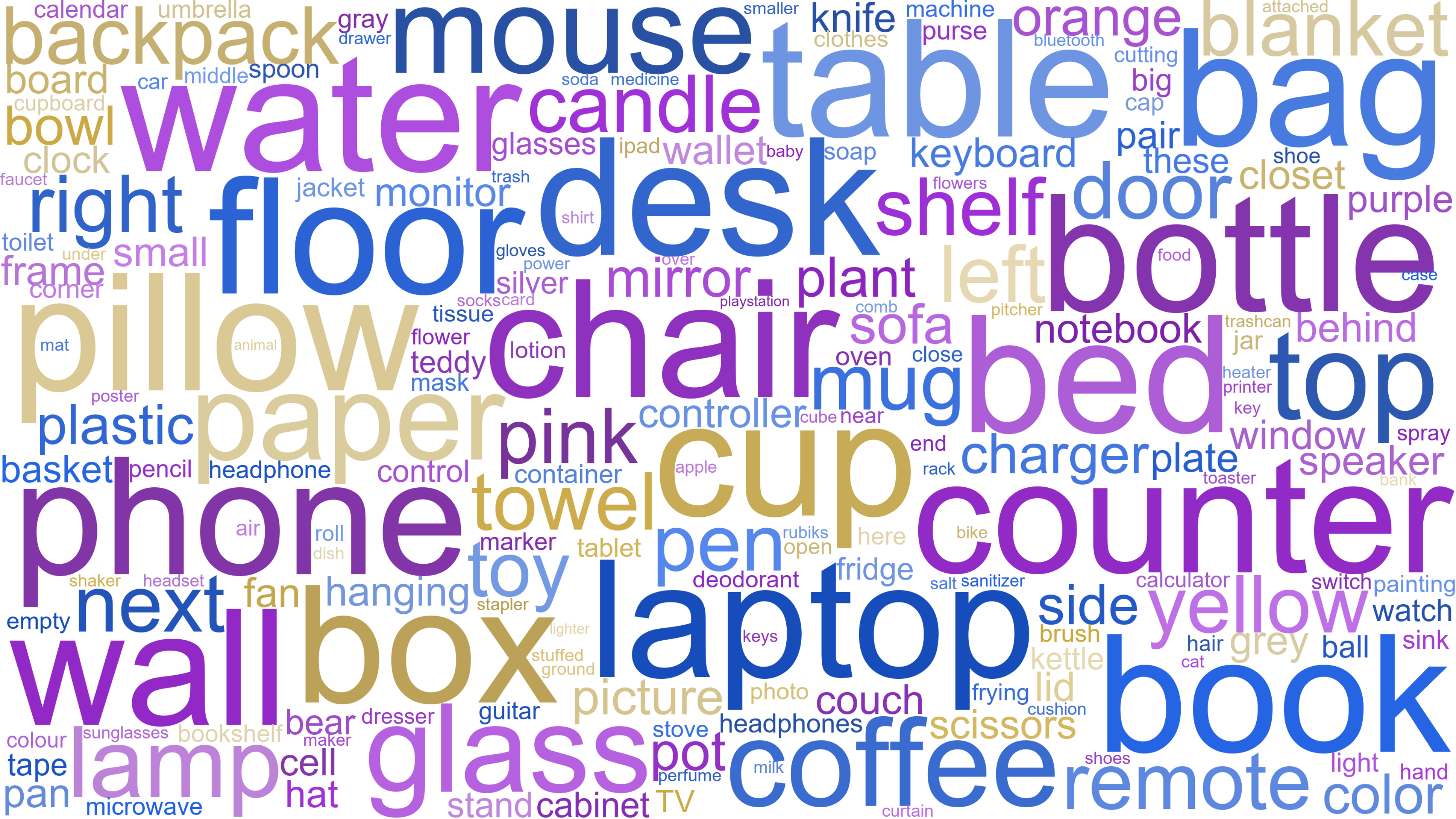}
    \caption{}
\end{subfigure}
\caption{SIGAR dataset statistics. (a) the word cloud of state expression. (b) the word cloud of intention expression. (c) the word cloud of object category.}
\label{fig3}
\end{figure}

\begin{figure}[htbp]
\centering
\begin{subfigure}[b]{0.32\columnwidth}
    \includegraphics[width=\textwidth]{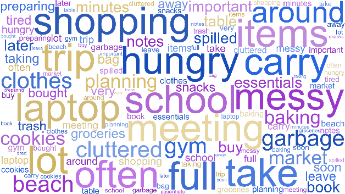}
    \caption{}
\end{subfigure}
\hfill
\begin{subfigure}[b]{0.32\columnwidth}
    \includegraphics[width=\textwidth]{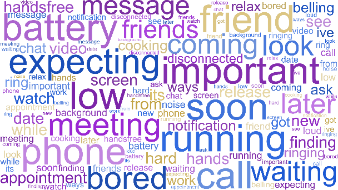}
    \caption{}
\end{subfigure}
\hfill
\begin{subfigure}[b]{0.32\columnwidth}
    \includegraphics[width=\textwidth]{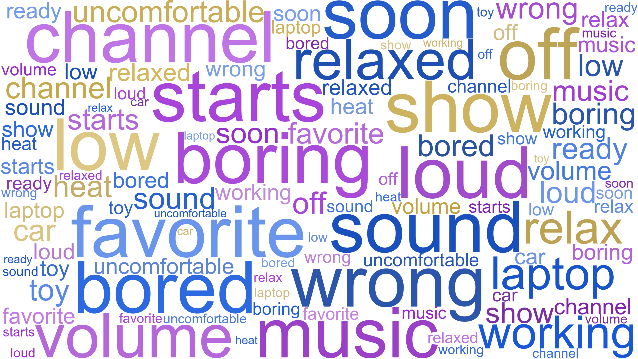}
    \caption{}
\end{subfigure}
\caption{State word clouds of partial categories from SIGAR dataset. (a) bag. (b) phone. (c) remote.}
\label{fig4}
\end{figure}

\begin{figure}[htbp]
\centering
\begin{subfigure}[b]{0.32\columnwidth}
    \includegraphics[width=\textwidth]{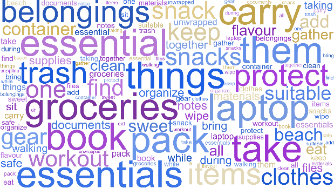}
    \caption{}
\end{subfigure}
\hfill
\begin{subfigure}[b]{0.32\columnwidth}
    \includegraphics[width=\textwidth]{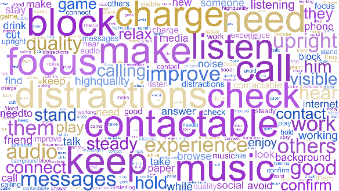}
    \caption{}
\end{subfigure}
\hfill
\begin{subfigure}[b]{0.32\columnwidth}
    \includegraphics[width=\textwidth]{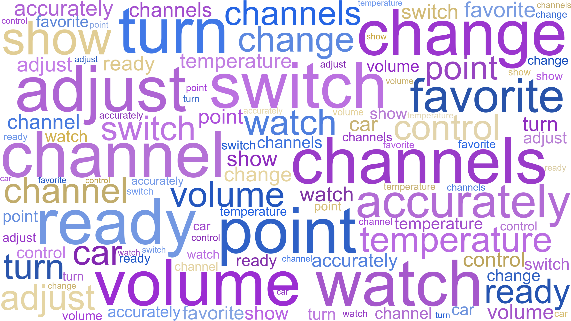}
    \caption{}
\end{subfigure}
\caption{Intention word clouds of partial categories from SIGAR dataset. (a) bag. (b) phone. (c) remote.}
\label{fig5}
\end{figure}

Our dataset comprises 4,195 images and 20,193 state \& intention expression texts. Fig. \ref{fig3} to Fig. \ref{fig5} visualize word clouds of different expression attributes. TABLE \ref{tab:tab1} compares our dataset with existing datasets, highlighting its distinctive features. While traditional datasets focus on direct object category descriptions, and both Affordance and intention-based datasets emphasize direct object-related references, our dataset uniquely captures:

\begin{itemize}
\item Indirect object-related state expressions
\item Human intention expressions
\item Embodied reference expression
\end{itemize}

\begin{table*}[htbp]
\caption{Comparison with Classic VG, AD and Intention-based Datasets}
\begin{center}
\begin{tabular}{|c|c|c|c|c|c|c|c|c|}
\hline
\multirow{2}*{\textbf{Datasets}} & \multirow{2}*{\textbf{Imgs}} & \multicolumn{4}{|c|}{\textbf{Labels}} & \multirow{2}*{\textbf{Gesture}} & \multirow{2}*{\textbf{Category}} & \multirow{2}*{\textbf{Avg Len}} \\ 
\cline{3-6}
& & \textbf{Num} & \textbf{State} & \textbf{Intention} & \textbf{Object} & & & \\ 
\hline
\multicolumn{9}{|l|}{\textbf{Classic VG}} \\
\hline
ReferIt \cite{kazemzadeh2014referitgame} & 20K & 97K & $\times$ & $\times$ & Phrases & $\times$ & 238 & 3.2 \\ 

RefCOCO \cite{kazemzadeh2014referitgame} & 20K & 50K & $\times$ & $\times$ & Phrases & $\times$ & 80 & 3.6 \\ 

RefCOCO+ \cite{kazemzadeh2014referitgame} & 20K & 49K & $\times$ & $\times$ & Phrases & $\times$ & 80 & 3.5 \\ 

RefCOCOg \cite{kazemzadeh2014referitgame} & 20K & 54K & $\times$ & $\times$ & Free-Form & $\times$ & 80 & 8.4 \\ 

GRES \cite{liu2023gres} & 20K & 60K & $\times$ & $\times$ & Phrases & $\times$ & 80 & 3.7 \\ 
\hline
\multicolumn{9}{|l|}{\textbf{Affordance Detection}} \\
\hline
ADE-Aff \cite{chuang2018learning} & 10K & 26K & $\times$ & Verbs & - & $\times$ & 150 & / \\ 

PAD \cite{luo2021one} & 4K & 4K & $\times$ & Verbs & - & $\times$ & 72 & / \\ 

PADV2 \cite{zhai2022one} & 30K & 30K & $\times$ & Verbs & - & $\times$ & 103 & / \\ 

COCO-Tasks \cite{sawatzky2019object} & 40K & 64K & $\times$ & Phrases & - & $\times$ & 49 & 2.6 \\ 
\hline
\multicolumn{9}{|l|}{\textbf{Intention-based VG}} \\
\hline
RIO \cite{qu2024rio} & 40K & 130K & $\times$ & Template & - & $\times$ & 69 & 15.7 \\ 

IntentionVG \cite{wang2024beyond}  & 100K & 500K & $\times$ & Free-Form & - & $\times$ & 1096 & 11.2 \\ 
\hline
\multicolumn{9}{|l|}{\textbf{Multi-attribute based VG}} \\
\hline
SIGAR & 4K & 20K & Free-Form & Free-Form & Free-Form & \checkmark & 395 & 17.1 \\ 
\hline
\multicolumn{9}{l}{Where Avg Len denote the average expression length. "-", "/" denote the expressions are unavailable.}
\end{tabular}
\label{tab:tab1}
\end{center}
\end{table*}

\subsection{Evaluation Metrics}
For the Multi-ref EC task, which combines state-intention texts and embodied references for object localization, we adopt standard Intersection over Union (IoU) thresholds at 0.25, 0.5, and 0.75. A prediction is considered successful when the IoU between predicted and ground truth bounding boxes exceeds the specified threshold, validating both localization accuracy and reference information inference.

\section{Baseline construction}

To address the Multi-ref EC task incorporating state-intention texts and embodied references, we propose two baseline approaches: an end-to-end approach and a model combination approach.

\subsection{End-to-end Models}
The end-to-end approach treats Multi-ref EC as a specialized REC task, where combined state-intention texts and images are processed as (image, text) pairs to derive object grounding information. We implement three categories of end-to-end baselines. First, we employ SOTA REC models designed for classical REC and VG tasks, including MDETR \cite{kamath2021mdetr}, SeqTR \cite{zhu2022seqtr}, GLIP \cite{li2022grounded}, Grounding DINO \cite{liu2023grounding}, and TOIST \cite{li2022toist}. Second, we utilize the ERU model, a specialized grounding method developed for the YouRefIt dataset, which utilizes touch-line Transformer to process both textual cues and agent gestures for improved spatial localization accuracy.

For the third category, we implement an MLLM approach using Qwen-VL, chosen for its robust scenario understanding and reasoning capabilities. We carefully design a set of prompts that enable efficient processing of state, intention, and embodied references. These prompts are structured to work effectively whether the different types of references are used independently or in combination to accomplish the Multi-ref EC task.

\subsection{Model Combination Approaches}
The model combination approach employs a two-step pipeline that integrates MLLM with end-to-end models. In this approach, we use Qwen2-VL \cite{Qwen2VL} as a front-end text interpreter to perceive and map state-intention texts to corresponding objects through comprehensive scene understanding. The text interpreter analyzes the scene and identifies object categories that align with the given states and intentions.

Following the text interpretation phase, the identified object categories are fed into an end-to-end REC model for precise object localization. To ensure optimal performance, both the MLLM text interpreter and the REC model undergo separate fine-tuning before integration. This combination leverages the MLLM's strong scene perception and inference capabilities while maintaining accurate object localization through proven REC models.

\section{Experiment}

\subsection{Implementation Details}

For the model combination approach, we fine-tune the text interpreter to output target object categories and train the end-to-end model with ground truth objective categories as input. The two components are trained separately before integration. For the end-to-end approach, we directly train the baselines using state and intention texts as input. All models are initialized with their original pre-trained checkpoints, maintaining their default hyperparameters except for minor batch size adjustments based on our GPU constraints. All experiments are conducted on NVIDIA 4090 GPUs.

\subsection{Results Comparison and Analysis}

We evaluated visual grounding performance for the Multi-ref EC task using a newly constructed SIGAR test set, focusing on state-intent combined with gestural information. As shown in TABLE \ref{tab:tab2}, baseline models were categorized into combinatorial methods and end-to-end models. We re-implemented and evaluated these SOTA methods on our SIGAR dataset to ensure fair comparison.

\begin{table}[tb]
\caption{Comparisons with the Classic VG SOTA Approaches on SIGAR Dataset}
\label{tab:tab2}
\begin{center}
\begin{tabular}{|l|c|c|c|}
\hline
\multirow{2}{*}{\textbf{Baselines}} & \multicolumn{3}{c|}{\textbf{IoU Threshold}} \\
\cline{2-4}
& \textbf{0.25} & \textbf{0.5} & \textbf{0.75} \\
\hline
\multicolumn{4}{|l|}{\textbf{Combination}} \\
\hline
Interpreter + Grounding-DINO & 39.6 & 35.6 & 29.2 \\
Interpreter + GLIP & 22.2 & 15.8 & 10.2 \\
Interpreter + touch-line Transformer & 26.1 & 21.7 & 12.5 \\
\hline
\multicolumn{4}{|l|}{\textbf{End-to-end}} \\
\hline
MDETR \cite{kamath2021mdetr} & 42.5 & 35.3 & 30.5 \\
TOIST \cite{li2022toist} & 43.2 & 40.4 & 32.8 \\
SeqTR \cite{zhu2022seqtr} & 44.0 & 42.5 & 35.1 \\
Grounding-DINO \cite{liu2023grounding} & 45.6 & 40.9 & 26.6 \\
GLIP \cite{li2022grounded} & 34.9 & 29.6 & 26.1 \\
touch-line Transformer \cite{li2023understanding} & 53.7 & 46.4 & 32.1 \\
\hline
\multicolumn{4}{|l|}{\textbf{MLLM}} \\
\hline
Qwen-VL \cite{Qwen-VL} & \textbf{54.4} & \textbf{51.9} & \textbf{32.3} \\
\hline
\end{tabular}
\end{center}
\end{table}

The experimental results revealed several key findings. End-to-end models demonstrated superior performance compared to combinatorial approaches, with multimodal large language models outperforming specialized end-to-end VL models. This superiority can be attributed to two possible factors. First, large language models show inherent limitations when serving as translators in combinatorial approaches. While operating as GQA models for target object categorization, they demonstrate restricted ability in comprehending implicit information and processing complex semantic relationships, unlike their end-to-end counterparts Second, the fragmented optimization objectives in combinatorial models, where components are independently optimized, may achieve local optima at the expense of global performance. Additionally, error propagation through the processing pipeline leads to cumulative performance degradation.

Among the baselines, touch-line Transformer and Qwen-VL demonstrated notable advantages, primarily due to their superior embodied reference understanding capabilities. Touch-line Transformer achieves this through modeling collinearity between gesture keypoints and object centers, while Qwen-VL utilizes our designed prompt to enhance spatial localization accuracy. The relatively limited performance of touch-line Transformer in combinatorial approaches may be attributed to low collinearity between interpreter-generated object categories and gesture keypoints, creating conflicts during inference that constrain modeling capabilities.

\begin{figure*}[ht]
\centering
\begin{tabular}{ccc}

\textbf{Grounding-DINO} & \textbf{Interpreter + touch-line Transformer} & \textbf{Qwen-VL} \\
\includegraphics[width=0.3\textwidth]{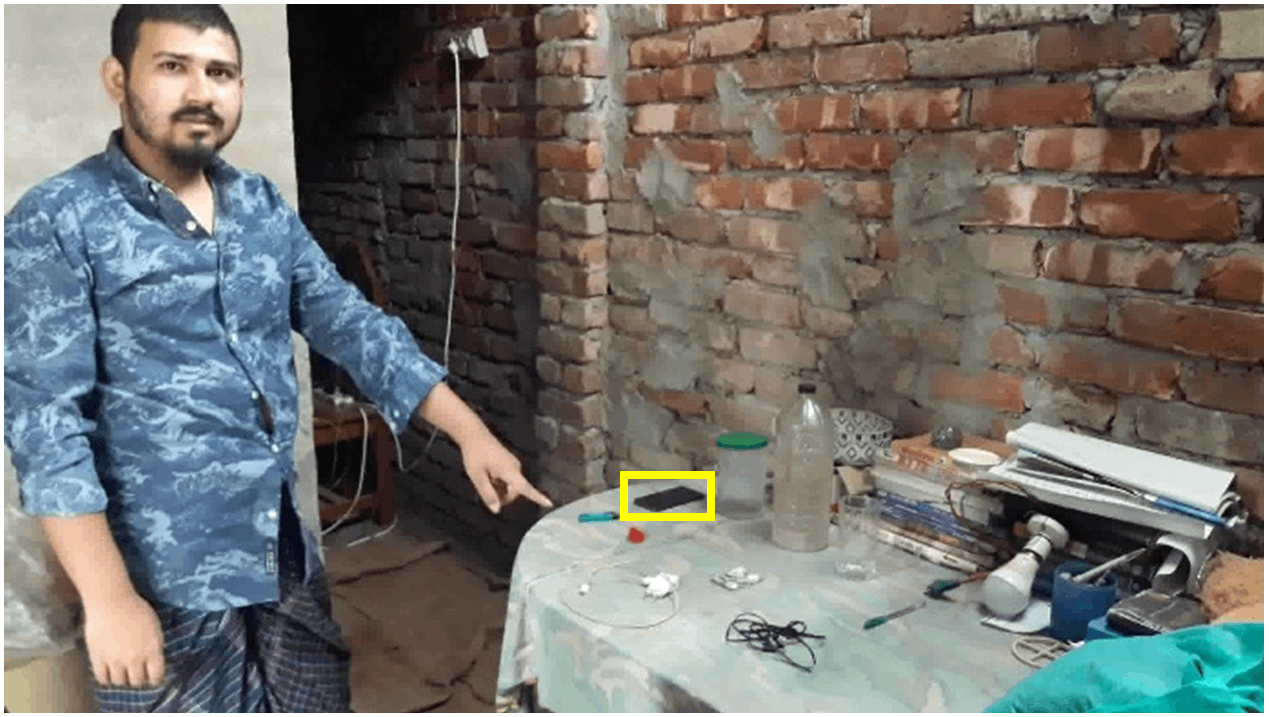} & \includegraphics[width=0.3\textwidth]{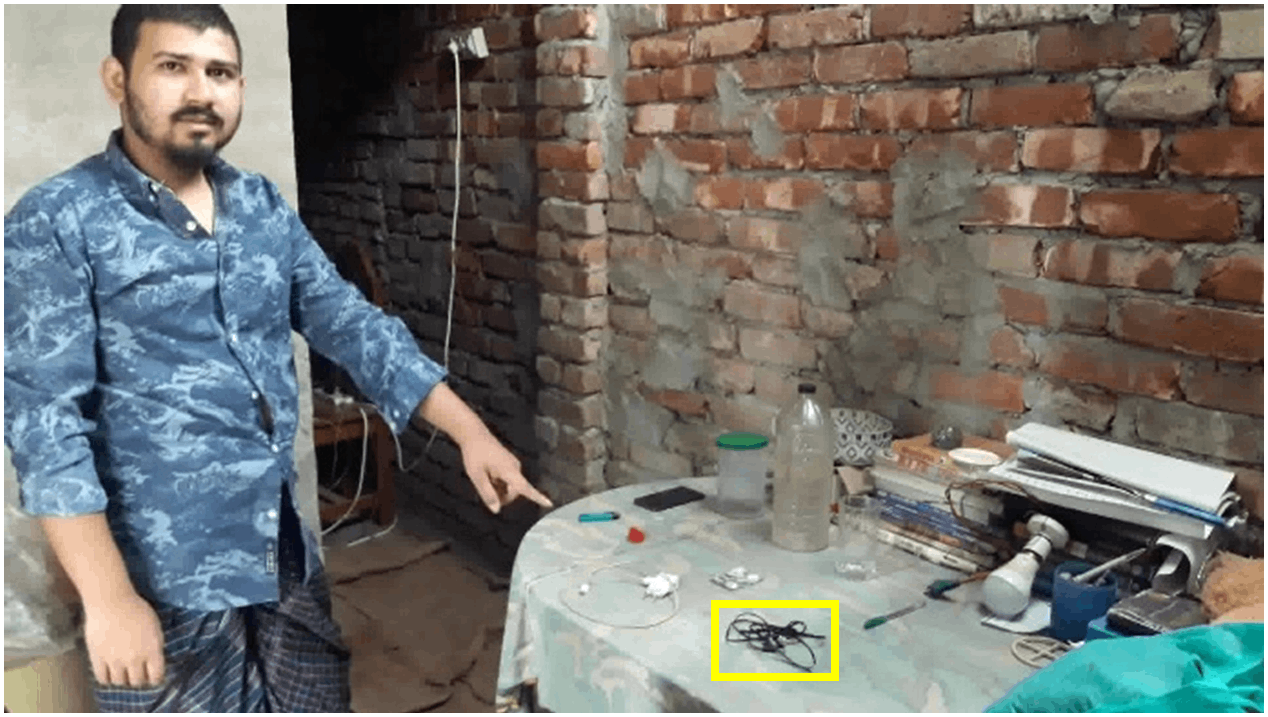} & \includegraphics[width=0.3\textwidth]{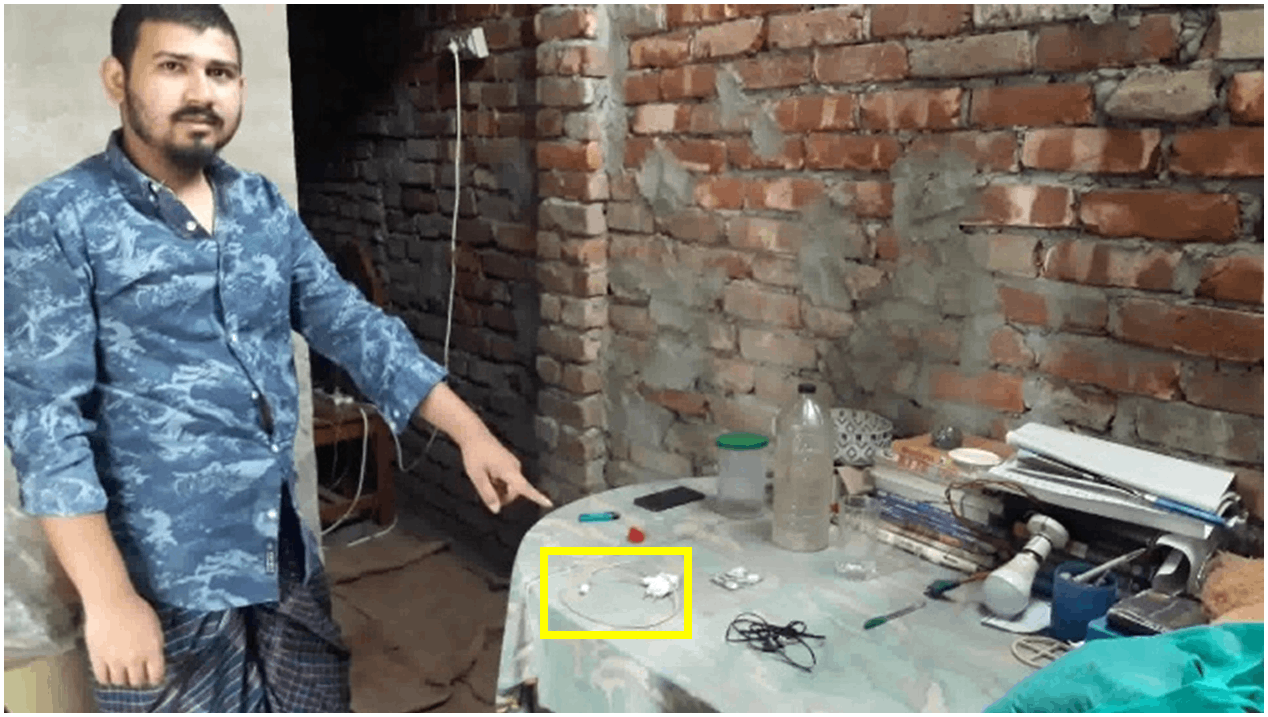} \\
\multicolumn{3}{l}{I need to charge the phone and my phone is running low on battery and focus where I am pointing} \vspace{10pt}\\

\includegraphics[width=0.3\textwidth]{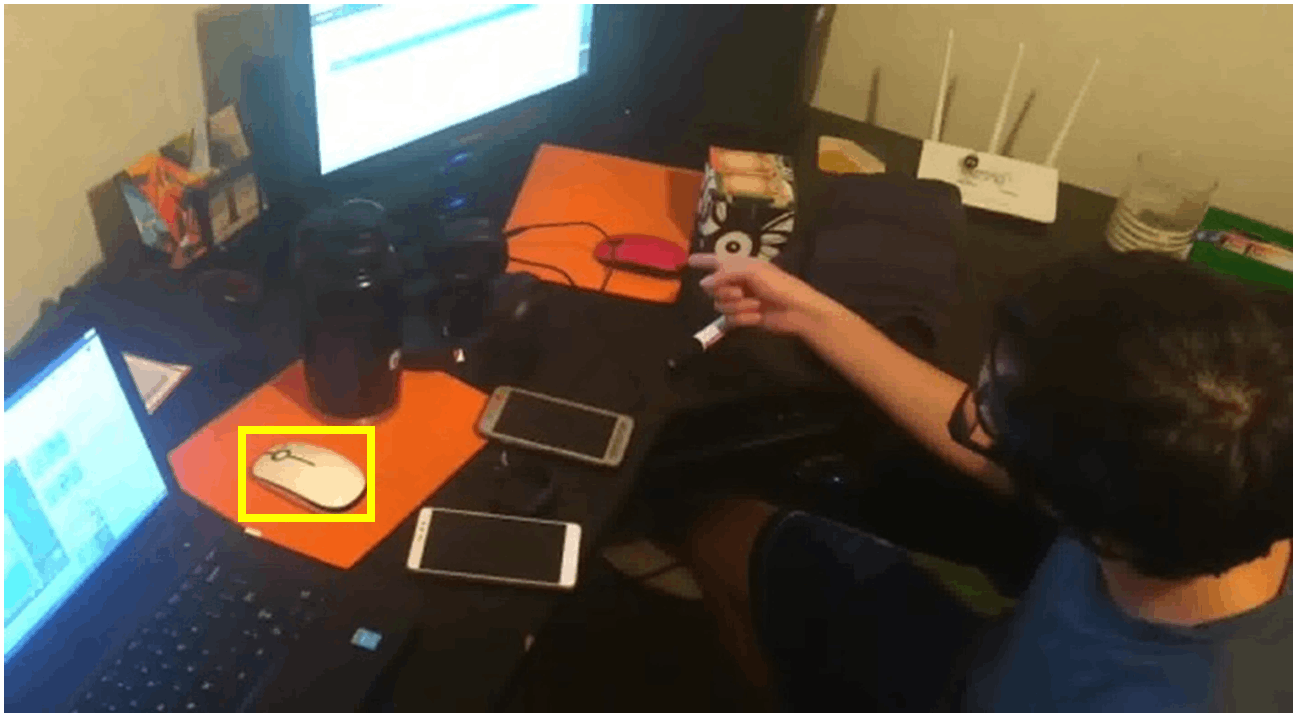} & \includegraphics[width=0.3\textwidth]{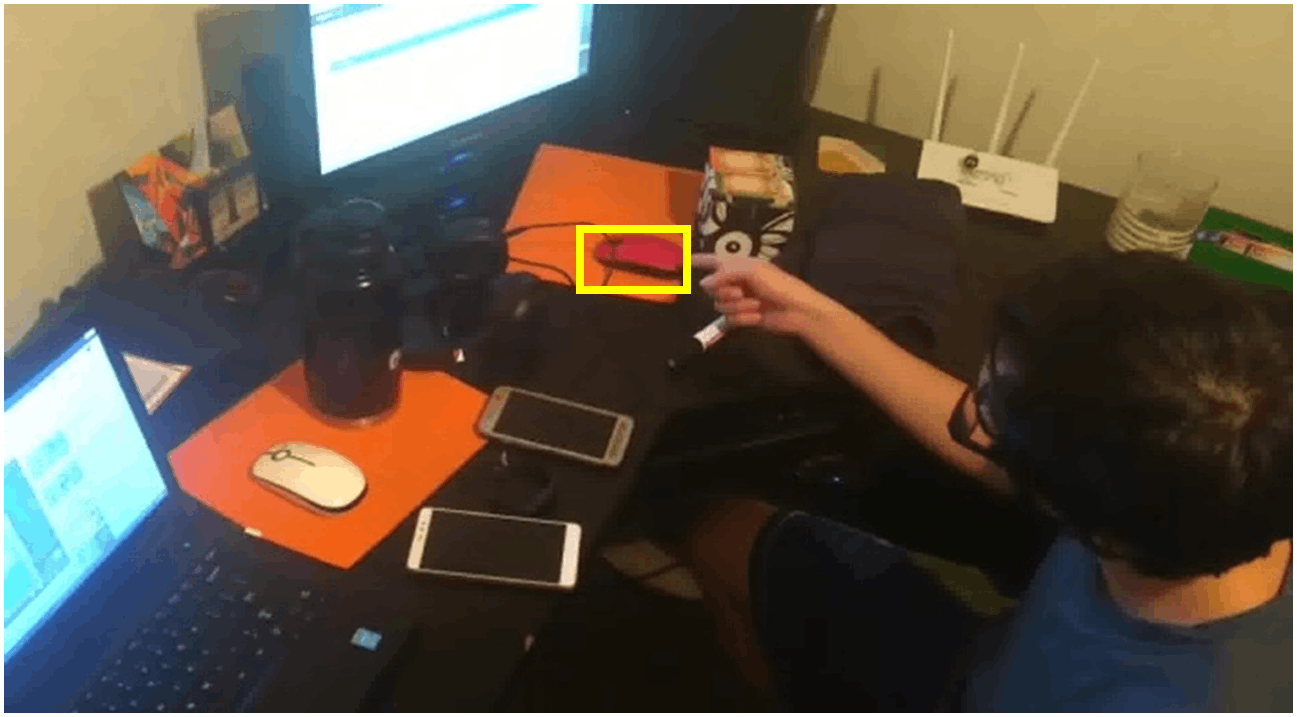} & \includegraphics[width=0.3\textwidth]{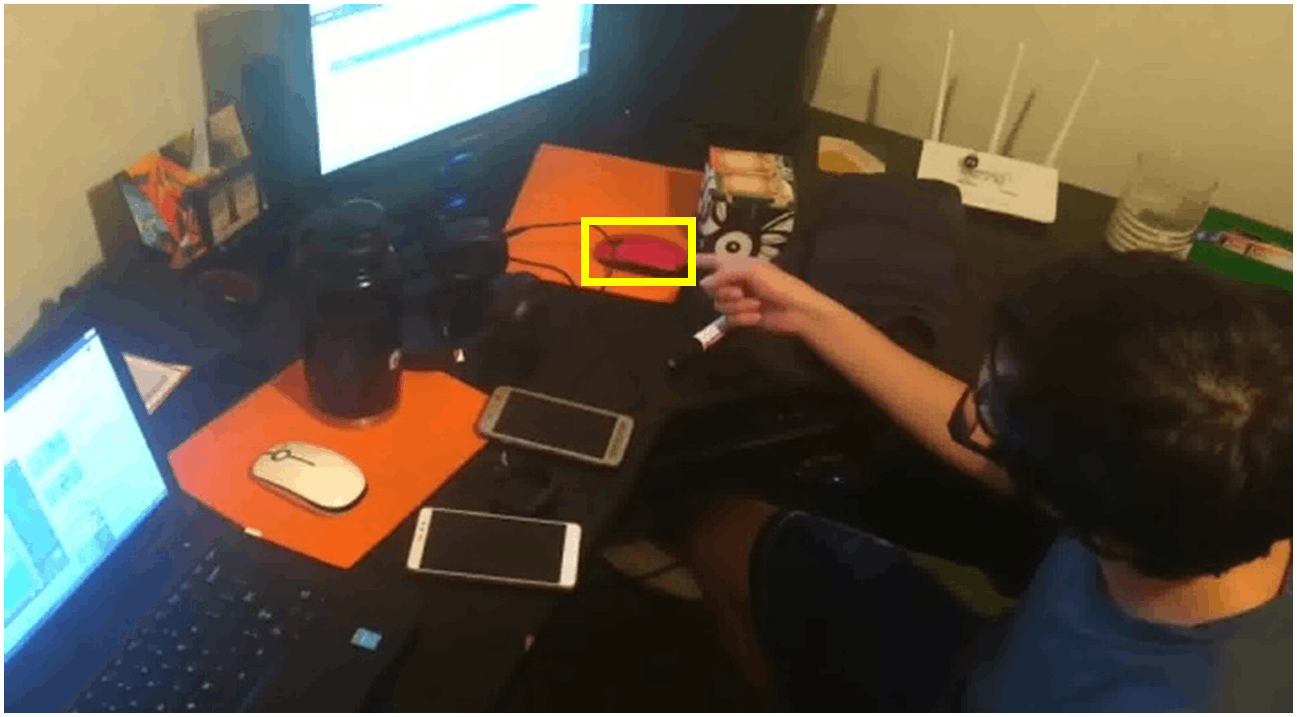} \\
\multicolumn{3}{l}{I need to point icons accurately and I am working on my computer and focus where I am pointing} \vspace{10pt}\\

\includegraphics[width=0.3\textwidth]{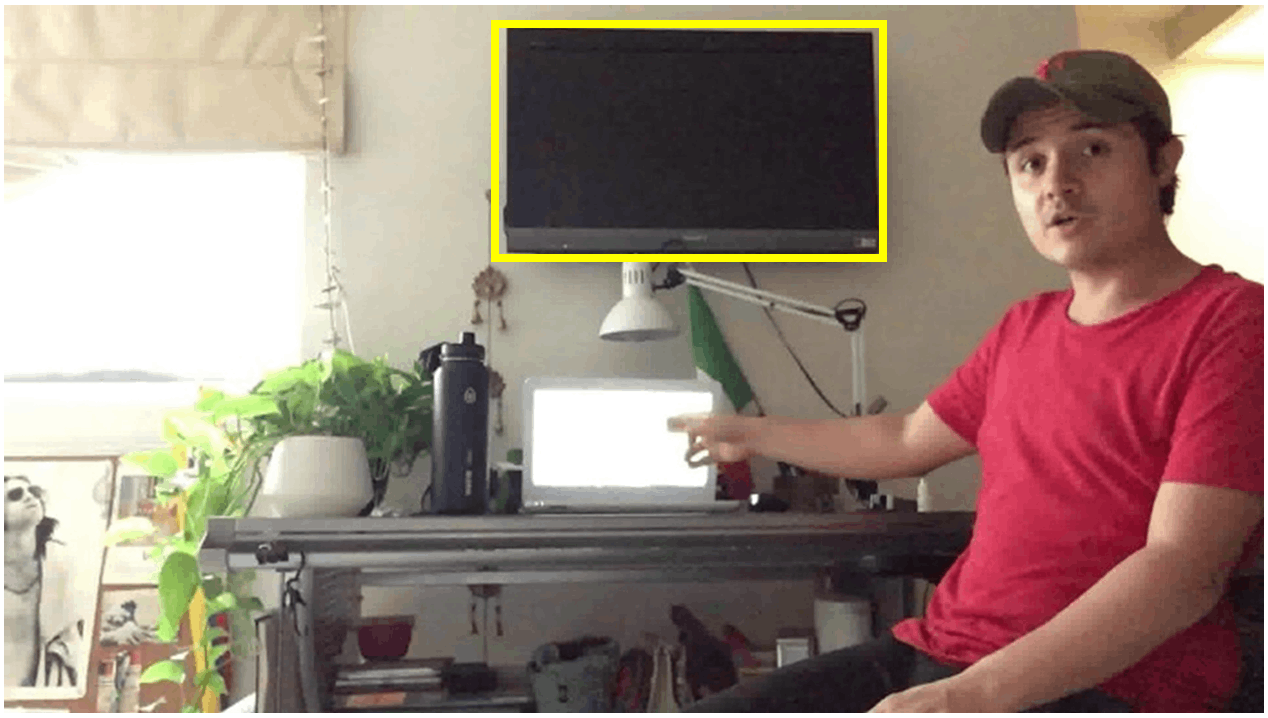} & \includegraphics[width=0.3\textwidth]{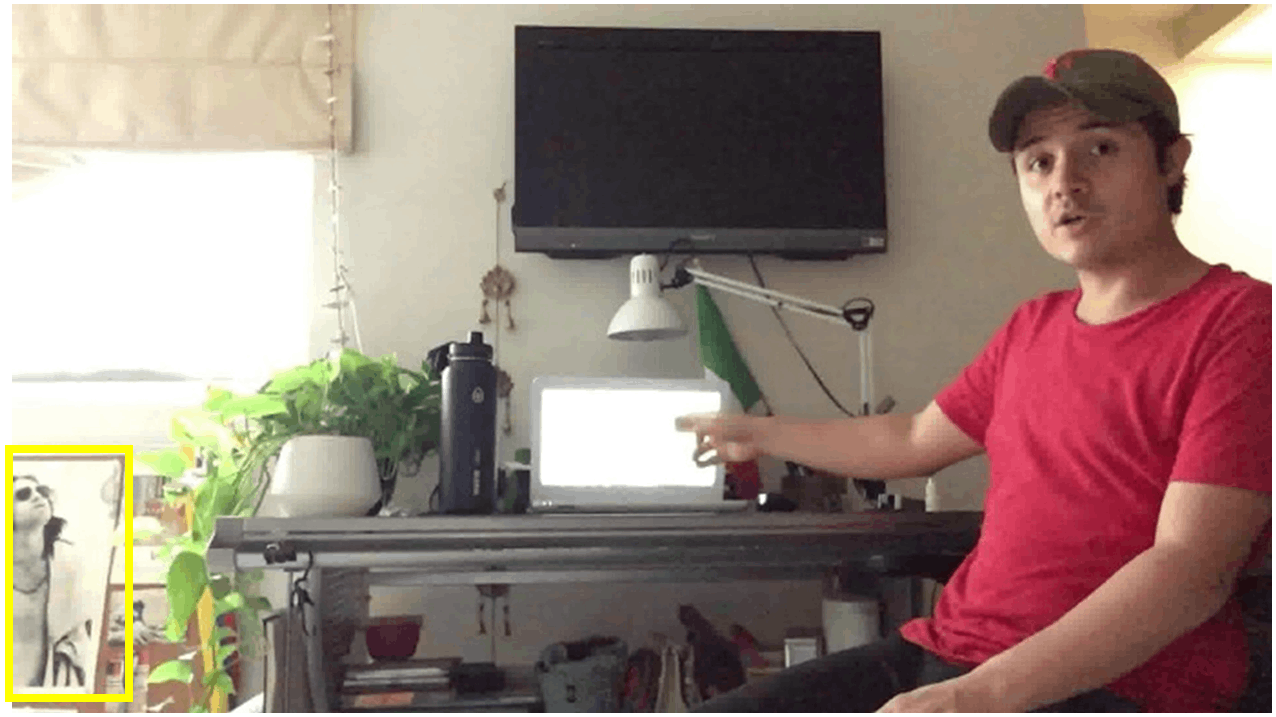} & \includegraphics[width=0.3\textwidth]{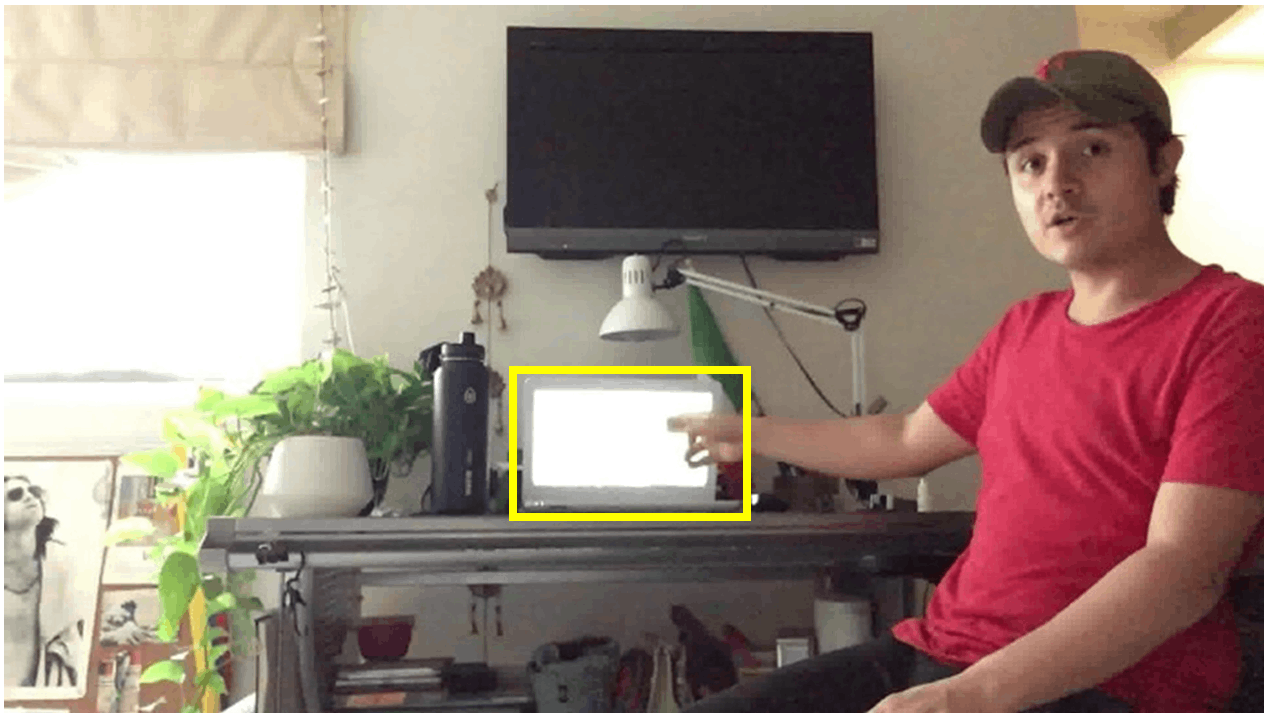} \\
\multicolumn{3}{l}{I want to watch a movie for rest and I just finished my work and focus where I am pointing} \\
\end{tabular}
\caption{The comparison of VG results between different baselines on SIGAR dataset.}
\label{fig:fig6}
\end{figure*}

Qualitative analysis in Fig. \ref{fig:fig6} shows that Qwen-VL, incorporating both state-intention texts and gesture understanding prompts, achieves better localization accuracy compared to both combination approaches (Interpreter + touch-line Transformer) and SOTA end-to-end models (Grounding-DINO).

\subsection{Ablation Study}
To validate the effectiveness and evaluate the contribution of different attribute modalities in visual grounding tasks, we conducted comprehensive ablation studies on the SIGAR dataset. We employed three IoU thresholds as evaluation metrics for object localization accuracy. Qwen-VL, which demonstrates superior capabilities in both general text comprehension and gesture understanding, served as our baseline model for these ablation experiments.
\paragraph{Single Attribute Reference Analysis}
We evaluated the individual impact of different reference attributes from the SIGAR dataset as standalone text inputs, with results presented in TABLE \ref{tab:tab3}. Among individual references, intention text demonstrated superior performance, followed by state text, while gesture text showed the lowest effectiveness. This performance variation may be attributed to the differences in semantic association strength. Intention text typically contains explicit action descriptions, establishing strong functional associations between actions and their corresponding object categories, thereby facilitating more effective intent-object mapping. In contrast, state text conveys conditions or implicit needs, requiring multi-step semantic reasoning: from state interpretation to action inference, and finally to object identification. This indirect semantic association pathway inherently increases the complexity of comprehension and reasoning.

\begin{table}[bt]
\centering
\caption{Comparisons of Single Attribute Reference Performance}
\begin{center}
\begin{tabular}{|l|c|c|c|}
\hline
\multirow{2}{*}{\textbf{Attribute}} & \multicolumn{3}{c|}{\textbf{IoU Threshold}} \\
\cline{2-4}
& \textbf{0.25} & \textbf{0.5} & \textbf{0.75} \\
\hline
State & 44.9 & 35.9 & 20.9 \\

Intention & 48.5 & 38.9 & 22.2 \\

Gesture & 35.4 & 27.4 & 17.9 \\
\hline
\end{tabular}
\end{center}
\label{tab:tab3}
\end{table}

\begin{table}[ht]  
\caption{Comparisons of Dual Attribute Reference Performance}
\begin{center}
\begin{tabular}{|l|c|c|c|}
\hline
\multirow{2}{*}{\textbf{Dual Attribute}} & \multicolumn{3}{c|}{\textbf{IoU Threshold}} \\
\cline{2-4}
& \textbf{0.25} & \textbf{0.5} & \textbf{0.75} \\
\hline
State + Intention & 51.1 & 45.1 & 26.1 \\

State + Gesture & 48.8 & 40.0 & 22.8 \\

Intention + Gesture & 53.2 & 47.3 & 28.1 \\
\hline
\end{tabular}
\end{center}
\label{tab:tab4}
\end{table}

The relatively lower performance of gesture text may be attributed to two primary factors. Firstly, gesture-based descriptions rely solely on spatial deictic relations. Without complementary semantic constraints, the model struggles to identify target objects among multiple candidates along the gestural vector, highlighting the limitations of pure spatial descriptions. Secondly, despite its robust general comprehension capabilities, the multimodal large language model exhibits limitations in processing precise spatial relationships, particularly when strict adherence to gestural covariance is required for object localization. These performance disparities reveal characteristic behaviors of multimodal large language models in processing diverse textual inputs: the model exhibits stronger performance with direct semantic associations while showing relative weakness in processing descriptions requiring complex spatial reasoning. These findings provide valuable insights for understanding the semantic processing mechanisms of multimodal large language models and optimizing input representations.

\begin{table*}[htbp]
\caption{Comparison of Different Attribute Orders in Prompts}
\begin{center}
\begin{tabular}{|p{10cm}|c|c|c|}  
\hline
\multirow{2}{*}{\textbf{Prompt}} & \multicolumn{3}{c|}{\textbf{IoU Threshold}} \\
\cline{2-4}
& \textbf{0.25} & \textbf{0.5} & \textbf{0.75} \\
\hline
\multicolumn{4}{|l|}{\textbf{Concentrated distribution with linguistic information at the front}} \\
\hline
\{grounding instruction\}$^{\mathrm{a}}$ \{intention\} and \{desire\}, and focus where I am pointing & 54.4 & 51.9 & 32.3 \\

\{grounding instruction\} \{desire\} and \{intention\}, and focus where I am pointing & 50.8 & 49.2 & 30.5 \\
\hline
\multicolumn{4}{|l|}{\textbf{Concentrated distribution with non-linguistic information at the front}} \\
\hline
\{grounding instruction\} I am pointing at, \{intention\} and \{desire\} & 36.6 & 37.3 & 23.0 \\

\{grounding instruction\} I am pointing at, \{desire\} and \{intention\} & 32.4 & 34.1 & 21.6 \\
\hline
\multicolumn{4}{|l|}{\textbf{Dispersed distribution}} \\
\hline
\{intention\} and \{desire\}, \{grounding instruction\} I am pointing at & 30.0 & 32.1 & 21.0 \\

\{desire\} and \{intention\}, \{grounding instruction\} I am pointing at & 29.1 & 31.7 & 20.9 \\
\hline
\multicolumn{4}{l}{$^{\mathrm{a}}$\{grounding instruction\}: output the bounding box of the object that} \\
\end{tabular}
\end{center}
\label{tab:tab5}
\end{table*}

\paragraph{Dual Attribute Reference Analysis}
Building upon single-attribute analysis, we investigated the performance of attribute combinations. As illustrated in TABLE \ref{tab:tab4}, among the three combinatorial approaches, intention + Gesture demonstrated optimal performance, followed by State + Intention, while State + Gesture showed the lowest effectiveness. This performance hierarchy remained consistent across all IoU thresholds. The superior performance of the Intention + Gesture combination highlights the complementary nature of these modalities. Intention text may provide semantic constraints that narrow down potential object categories, while gestural information offers precise spatial localization cues. This synergistic integration significantly enhances the model's visual grounding accuracy.

Although the State + Intention combination offers multi-perspective semantic descriptions of target objects, the semantic overlap between these modalities potentially limits their combinatorial effectiveness. The relatively lower performance of the State + Gesture combination may be attributed to the ambiguous semantic information from state text. This semantic ambiguity, coupled with spatial gestural cues, fails to provide sufficiently distinctive target object indicators. These findings elucidate the synergistic effects between different textual modalities and provide valuable insights for optimizing multimodal information integration strategies.

\paragraph{Prompt Component Order Analysis}
In order to study the effect of prompt Component Order on model performance, we designed six different text arrangements and divided them into three categories according to the text arrangement characteristics. prompt was designed with three main dimensions in mind to consider the effect on model prediction: the centralized and decentralized placement of descriptive information, the relative position of verbal and gesture information, and the relative position of the state information and intention information within the verbal information, as shown in TABLE \ref{tab:tab5}.

The experimental results reveal three important findings: first, by comparing the overall performance of the three types of distribution, it can be found that the centralized distribution of descriptive text is significantly better than the decentralized distribution. Although the dispersed distribution of descriptive text may correspond to utterances that are more in line with human natural language expression habits, the experimental results suggest that the large model may prefer a compact semantic organization when processing structured information. Second, by comparing the distribution of the two types of text concentration with linguistic information in front and gesture information in front, it can be clearly seen that the prompt format that puts the linguistic information before the gesture information has a significant advantage, which is in line with the human cognitive habit of ordering, i.e., to understand the semantic features of the target object first, and then to determine its spatial location, which enables the model to process the multimodal information more efficiently. Finally, a comparison of the relative positions of intention and state reveals that model performance is consistently better when intention information is located before state information. This phenomenon was consistently verified across all groups. This suggests that the explicit behavioral descriptions provided by the intention text help establish a clearer semantic framework and provide a better context for the subsequent integration of the state information. This finding not only deepens our understanding of the importance of text order in prompt engineering, but also reveals possible differences in language understanding between large models and human cognition.

\section{Conclusion}
This paper introduces Multi-ref EC, a novel visual grounding task that extends beyond traditional object category descriptions by integrating state expressions, intention expressions, and embodied references in HRI scenarios. Our approach recognizes that users express desires through multiple attribute references, requiring agents to comprehend and integrate these references for accurate object localization. To facilitate research in this direction, we present SIGAR, a comprehensive dataset featuring free-form expressions of states, intentions, and gesture references. Through extensive experiments with various baseline models, we demonstrate that properly ordered multi-attribute references contribute to improved visual grounding performance. Our analysis reveals the necessity of multi-attribute references and their optimal organization patterns for effective visual grounding. This work opens new research directions in multi-attribute reference understanding, with SIGAR serving as a valuable benchmark for exploring more natural human-robot interaction.


\bibliographystyle{ieeetr}
\bibliography{IEEEabrv, references}

\begin{thebibliography}{10}

\bibitem{tang2020bootstrapping}
N.~Tang, S.~Stacy, M.~Zhao, G.~Marquez, and T.~Gao, ``Bootstrapping an imagined we for cooperation.,'' in {\em CogSci}, 2020.

\bibitem{herbort2016spatial}
O.~Herbort and W.~Kunde, ``Spatial (mis-) interpretation of pointing gestures to distal referents.,'' {\em Journal of Experimental Psychology: Human Perception and Performance}, vol.~42, no.~1, p.~78, 2016.

\bibitem{herbort2018point}
O.~Herbort and W.~Kunde, ``How to point and to interpret pointing gestures? instructions can reduce pointer--observer misunderstandings,'' {\em Psychological research}, vol.~82, no.~2, pp.~395--406, 2018.

\bibitem{qu2024rio}
M.~Qu, Y.~Wu, W.~Liu, X.~Liang, J.~Song, Y.~Zhao, and Y.~Wei, ``Rio: A benchmark for reasoning intention-oriented objects in open environments,'' {\em Advances in Neural Information Processing Systems}, vol.~36, 2024.

\bibitem{wang2024beyond}
W.~Wang, Y.~Zhang, X.~He, Y.~Yan, Z.~Zhao, X.~Wang, and J.~Liu, ``Beyond literal descriptions: Understanding and locating open-world objects aligned with human intentions,'' {\em arXiv preprint arXiv:2402.11265}, 2024.

\bibitem{qiu2020human}
S.~Qiu, H.~Liu, Z.~Zhang, Y.~Zhu, and S.-C. Zhu, ``Human-robot interaction in a shared augmented reality workspace,'' in {\em 2020 IEEE/RSJ International Conference on Intelligent Robots and Systems (IROS)}, pp.~11413--11418, IEEE, 2020.

\bibitem{li2023understanding}
Y.~Li, X.~Chen, H.~Zhao, J.~Gong, G.~Zhou, F.~Rossano, and Y.~Zhu, ``Understanding embodied reference with touch-line transformer.,'' in {\em ICLR}, 2023.

\bibitem{chen2021yourefit}
Y.~Chen, Q.~Li, D.~Kong, Y.~L. Kei, S.-C. Zhu, T.~Gao, Y.~Zhu, and S.~Huang, ``Yourefit: Embodied reference understanding with language and gesture. 2021 ieee,'' in {\em CVF International Conference on Computer Vision (ICCV)}, 2021.

\bibitem{zhang2024vision}
J.~Zhang, J.~Huang, S.~Jin, and S.~Lu, ``Vision-language models for vision tasks: A survey,'' {\em IEEE Transactions on Pattern Analysis and Machine Intelligence}, 2024.

\bibitem{yao2024detclipv3}
L.~Yao, R.~Pi, J.~Han, X.~Liang, H.~Xu, W.~Zhang, Z.~Li, and D.~Xu, ``Detclipv3: Towards versatile generative open-vocabulary object detection,'' in {\em Proceedings of the IEEE/CVF Conference on Computer Vision and Pattern Recognition}, pp.~27391--27401, 2024.

\bibitem{kamath2021mdetr}
A.~Kamath, M.~Singh, Y.~LeCun, G.~Synnaeve, I.~Misra, and N.~Carion, ``Mdetr-modulated detection for end-to-end multi-modal understanding,'' in {\em Proceedings of the IEEE/CVF international conference on computer vision}, pp.~1780--1790, 2021.

\bibitem{li2022grounded}
L.~H. Li, P.~Zhang, H.~Zhang, J.~Yang, C.~Li, Y.~Zhong, L.~Wang, L.~Yuan, L.~Zhang, J.-N. Hwang, {\em et~al.}, ``Grounded language-image pre-training,'' in {\em Proceedings of the IEEE/CVF Conference on Computer Vision and Pattern Recognition}, pp.~10965--10975, 2022.

\bibitem{liu2023grounding}
S.~Liu, Z.~Zeng, T.~Ren, F.~Li, H.~Zhang, J.~Yang, C.~Li, J.~Yang, H.~Su, J.~Zhu, {\em et~al.}, ``Grounding dino: Marrying dino with grounded pre-training for open-set object detection,'' {\em arXiv preprint arXiv:2303.05499}, 2023.

\bibitem{pi2023detgpt}
R.~Pi, J.~Gao, S.~Diao, R.~Pan, H.~Dong, J.~Zhang, L.~Yao, J.~Han, H.~Xu, L.~Kong, {\em et~al.}, ``Detgpt: Detect what you need via reasoning,'' {\em arXiv preprint arXiv:2305.14167}, 2023.

\bibitem{claude2024sonnet}
Anthropic, ``Claude model - anthropic,'' {\em URL https://www.anthropic.com/claude}, Accessed: 2024-03.

\bibitem{kazemzadeh2014referitgame}
S.~Kazemzadeh, V.~Ordonez, M.~Matten, and T.~Berg, ``Referitgame: Referring to objects in photographs of natural scenes,'' in {\em Proceedings of the 2014 conference on empirical methods in natural language processing (EMNLP)}, pp.~787--798, 2014.

\bibitem{liu2023gres}
C.~Liu, H.~Ding, and X.~Jiang, ``Gres: Generalized referring expression segmentation,'' in {\em Proceedings of the IEEE/CVF conference on computer vision and pattern recognition}, pp.~23592--23601, 2023.

\bibitem{chuang2018learning}
C.-Y. Chuang, J.~Li, A.~Torralba, and S.~Fidler, ``Learning to act properly: Predicting and explaining affordances from images,'' in {\em Proceedings of the IEEE Conference on Computer Vision and Pattern Recognition}, pp.~975--983, 2018.

\bibitem{luo2021one}
H.~Luo, W.~Zhai, J.~Zhang, Y.~Cao, and D.~Tao, ``One-shot affordance detection,'' {\em arXiv preprint arXiv:2106.14747}, 2021.

\bibitem{zhai2022one}
W.~Zhai, H.~Luo, J.~Zhang, Y.~Cao, and D.~Tao, ``One-shot object affordance detection in the wild,'' {\em International Journal of Computer Vision}, vol.~130, no.~10, pp.~2472--2500, 2022.

\bibitem{sawatzky2019object}
J.~Sawatzky, Y.~Souri, C.~Grund, and J.~Gall, ``What object should i use?-task driven object detection,'' in {\em Proceedings of the IEEE/CVF Conference on Computer Vision and Pattern Recognition}, pp.~7605--7614, 2019.

\bibitem{zhu2022seqtr}
C.~Zhu, Y.~Zhou, Y.~Shen, G.~Luo, X.~Pan, M.~Lin, C.~Chen, L.~Cao, X.~Sun, and R.~Ji, ``Seqtr: A simple yet universal network for visual grounding,'' in {\em European Conference on Computer Vision}, pp.~598--615, Springer, 2022.

\bibitem{li2022toist}
P.~Li, B.~Tian, Y.~Shi, X.~Chen, H.~Zhao, G.~Zhou, and Y.-Q. Zhang, ``Toist: Task oriented instance segmentation transformer with noun-pronoun distillation,'' {\em Advances in Neural Information Processing Systems}, vol.~35, pp.~17597--17611, 2022.

\bibitem{Qwen2VL}
P.~Wang, S.~Bai, S.~Tan, S.~Wang, Z.~Fan, J.~Bai, K.~Chen, X.~Liu, J.~Wang, W.~Ge, Y.~Fan, K.~Dang, M.~Du, X.~Ren, R.~Men, D.~Liu, C.~Zhou, J.~Zhou, and J.~Lin, ``Qwen2-vl: Enhancing vision-language model's perception of the world at any resolution,'' {\em arXiv preprint arXiv:2409.12191}, 2024.

\bibitem{Qwen-VL}
J.~Bai, S.~Bai, S.~Yang, S.~Wang, S.~Tan, P.~Wang, J.~Lin, C.~Zhou, and J.~Zhou, ``Qwen-vl: A versatile vision-language model for understanding, localization, text reading, and beyond,'' {\em arXiv preprint arXiv:2308.12966}, 2023.

\end{thebibliography}

\end{document}